\documentclass{article}

\usepackage{arxiv}

\usepackage[utf8]{inputenc} 
\usepackage[T1]{fontenc}    
\usepackage{hyperref}       
\usepackage{url}            
\usepackage{booktabs}       
\usepackage{amsfonts}       
\usepackage{nicefrac}       
\usepackage{microtype}      
\usepackage{lipsum}		
\usepackage{graphicx}
\usepackage{natbib}
\usepackage{doi}

\usepackage{xcolor}
\hypersetup{
    colorlinks, 
    linkcolor={red!50!black},
    citecolor={blue!50!black},
    urlcolor={blue!80!black} 
}

\title{Fast and Slow Planning}

\author{ \href{https://orcid.org/0000-0002-1161-0336}{\includegraphics[scale=0.06]{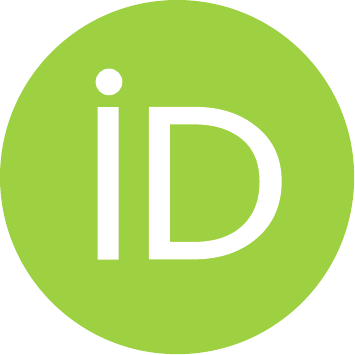}\hspace{1mm}Francesco Fabiano} \\
	University of Parma\\
	\texttt{francesco.fabiano@unipr.it} \\
	\And
        Vishal Pallagani \\
	University of South Carolina\\
	\texttt{vishalp@mailbox.sc.edu} \\
 	\And
	\href{https://orcid.org/0000-0001-6652-9853}{\includegraphics[scale=0.06]{orcid.pdf}\hspace{1mm}Marianna Bergamaschi Ganapini} \\
	Union College\\
	\texttt{bergamam@union.edu} \\
  	\And
	\href{https://orcid.org/0000-0001-6350-0238}{\includegraphics[scale=0.06]{orcid.pdf}\hspace{1mm}Lior Horesh} \\
	IBM Research\\
	\texttt{lhoresh@us.ibm.com} \\
  	\And
	\href{https://orcid.org/0000-0002-9846-0157}{\includegraphics[scale=0.06]{orcid.pdf}\hspace{1mm}Andrea Loreggia} \\
	University of Brescia\\
	\texttt{andrea.loreggia@unibs.it} \\
  	\And
	\href{https://orcid.org/0000-0001-6847-522X}{\includegraphics[scale=0.06]{orcid.pdf}\hspace{1mm}Keerthiram Murugesan} \\
	IBM Research\\
	\texttt{keerthiram.murugesan@ibm.com} \\
   	\And
	\href{https://orcid.org/0000-0001-8898-219X}{\includegraphics[scale=0.06]{orcid.pdf}\hspace{1mm}Francesca Rossi} \\
	IBM Research\\
	\texttt{francesca.rossi2@ibm.com} \\
    	\And
	\href{https://orcid.org/0000-0002-7292-3838}{\includegraphics[scale=0.06]{orcid.pdf}\hspace{1mm}Biplav Srivastava} \\
	University of South Carolina\\
	\texttt{biplav.s@sc.edu} \\
}

\renewcommand{\shorttitle}{Fast and Slow Planning}

\hypersetup{
pdftitle={Fast and Slow Planning},
pdfsubject={cs.AI},
pdfauthor={Francesco Fabiano, Vishal Pallagani, Marianna Bergamaschi Ganapini, Lior Horesh, Andrea Loreggia, Keerthiram Murugesan, Francesca Rossi, Biplav Srivastava}
}

\usepackage{xspace}
\usepackage[all]{foreign}
\usepackage[inline]{enumitem}
\usepackage{hhline}
\usepackage{algorithm}
\usepackage{algpseudocode}

\newcommand{\myvirgolette}[1]{``#1''}
\newcommand{\mep}{MEP}
\newcommand*{\EFP}{\mbox{E{\footnotesize  FP} 2.0}\xspace}
\newcommand{\sysStyle}[1]{\texttt{#1}}
\newcommand{\reasoner}{{\sysStyle{Plan-SOFAI}}\xspace}
\newcommand{\innsys}{\sysStyle{System}}
\newcommand{\mysys}{\innsys\xspace}
\newcommand{\mysysS}{\innsys s\xspace}
\newcommand{\sysone}{\sysStyle{\innsys-1}\xspace}
\newcommand{\systwo}{\sysStyle{\innsys-2}\xspace}
\newcommand{\abbrvSysOne}{\sysStyle{S1}\xspace}
\newcommand{\abbrvSysTwo}{\sysStyle{S2}\xspace}

\newcommand{\knobel}{knowledge\xspace}

\newcommand{\varstyle}[1]{\textit{#1}}
\newcommand{\funcstyle}[1]{\ensuremath{\mathtt{#1}}}
\newcommand{\MC}{\textbf{MC}}
\newcommand{\MCone}{\MC\textbf{-1}\xspace}
\newcommand{\MCtwo}{\MC\textbf{-2}\xspace}

\begin{document}
\maketitle

\begin{abstract}
The concept of Artificial Intelligence has gained a lot of attention over the last decade.
In particular, AI-based tools have been employed in several scenarios and are, by now, pervading our everyday life.
Nonetheless, most of these systems lack many capabilities that we would naturally consider to be included in a notion of ``intelligence''.
In this work, we present an architecture that, inspired by the cognitive theory known as \textit{Thinking Fast and Slow} by D. Kahneman, is tasked with solving planning problems in different settings, specifically: classical and multi-agent epistemic.
The system proposed is an instance of a more general AI paradigm, referred to as SOFAI (for Slow and Fast AI).
SOFAI exploits multiple solving approaches, with different capabilities that characterize them as either fast or slow, and a metacognitive module to regulate them.
This combination of components, which roughly reflects the human reasoning process according to D. Kahneman, allowed us to enhance the reasoning process that, in this case, is concerned with planning in two different settings.
The behavior of this system is then compared to state-of-the-art solvers, showing that the newly introduced system presents better results in terms of generality, solving a wider set of problems with an acceptable trade-off between solving times and solution accuracy.
\end{abstract}


\section{Introduction}
{
}

In the last few years, the AI community has produced several systems and techniques that allow solving autonomously and efficiently intricate problems of various natures.
These resolution procedures range from the use of formal logics to the creation of neural-based structures.
The former is an area of study that tries to define rational behavior for our systems while the latter, imitating the physiology of our brain, aims to emulate (to some degree) the human behavior.

In this work, we present and analyze an architecture that, exploiting several solving techniques, tackles planning problems in both the \emph{classical} and \emph{multi-agent epistemic} settings.
Our contribution is a specialization of the SOFAI architecture~\cite{booch-2021-thinkfast}, which, in turn, is inspired by the well-known cognitive theory \emph{Thinking Fast and Slow} by~\citet{kahneman2011thinking}.
The presented solving approach includes both ``fast'' and ``slow'' solvers and a metacognitive module to orchestrate the two.
Slow (or System 2, \systwo) solvers are solving problems by reasoning and (usually) exploiting symbolic techniques, while fast (or System 1, \sysone) solvers just employ past experience to identify the solution to a given problem.
Finally, the metacognitive module provides centralized governance and chooses the best solver for the problem at hand.

As already mentioned, in this paper, we consider the classical and multi-agent epistemic planning domains.
We employ an instance of the SOFAI architecture that includes
\begin{enumerate*}[label=\emph{\roman*)}]
\item existing planners as the slow solvers;
\item and both case-based plan selectors and the so-called \emph{Plansformer}~\cite{anonymous2023plansformer}, a Large Language Model (LLM) fine-tuned on planning problems which is capable of generating satisfying plans, as fast solvers.
\end{enumerate*}
Experimental results on widely used planning problem domains show that the behavior of our architecture is better than using the existing planner alone, both in terms of solved instances and average solving time.

Summarizing, the main contributions of this paper are:
\begin{itemize}
    \item The definition of a multi-agent architecture, inspired by the thinking fast and slow theory, to tackle classical and epistemic planning problems.
    \item The characterization of fast and slow solvers for the architecture, as well as the metacognitive module.
    \item Experimental results on planning domains, showing the behavior of the architecture when using different fast planners.
\end{itemize}

The paper is structured as follows.
After this brief introduction, we introduce the background knowledge on the thinking fast and slow theory and on classical and epistemic planning.
We then describe how to unify the main concepts of the thinking fast and slow theory in the planning environment, with special emphasis on the metacognitive module and the solvers. 
We follow with a description of the experimental setting, tables, and graphs providing the experimental results showing the behavior of our system on planning problems, and a discussion of such results.
We conclude the paper by summarizing the main contribution and hinting at ongoing work.

\section{Background}
\label{sec:background}

\subsection{Thinking Fast and Slow}

Thanks to improved algorithms, techniques, computational power, and dedicated hardware~\citep{marcus2020decade}, the various automated reasoning tools are becoming more and more efficient and reliable in dealing with their areas of interest.
However, all of these tools still lack capabilities that, we humans, naturally consider to be included in a notion of \myvirgolette{intelligence} as, for example, generalizability, robustness, and abstraction.
For these reasons, 
a growing segment of the AI community is attempting to address these limitations and is trying to create systems that display more \myvirgolette{human-like qualities}. 
One of the central strategies to tackle this problem, adopted by various research groups~\cite{newell1992soar,goel2017thinking,sofai2021}, 
envisions tools, usually referred to as cognitive architectures~\cite{Kotseruba2020}, that exploit a combination of both the aforementioned approaches.
In particular, in this paper, we explore classical and multi-agent epistemic planning in the context of one of these architectures that stems from a modern cognitive theory, \ie the well-known \emph{Thinking Fast and Slow} paradigm proposed by D. Kahneman~\cite{kahneman2011thinking}.

Kahneman's theory states that humans' reasoning capabilities are categorized into two diverse \myvirgolette{\mysysS}, called \sysone (\abbrvSysOne) and \systwo (\abbrvSysTwo).
In particular, \abbrvSysOne identifies the intuitive and imprecise decision-making processes (\myvirgolette{thinking fast}), while \abbrvSysTwo provides tools to tackle all those situations where complex decisions need to be taken following logical and rational thinking (\myvirgolette{thinking slow}).
Other than problem difficulty, \abbrvSysOne and \abbrvSysTwo discern which problem they should tackle based on the experience accumulated on the problem itself.
That is, when a \emph{new} non-trivial problem has to be solved, it is handled by \abbrvSysTwo.
However, certain problems that initially can be solved only by \abbrvSysTwo, can later be solved by \abbrvSysOne after having accumulated a certain amount of experience.
The reason is that the procedures used by \abbrvSysTwo to find solutions to such problems also accumulate examples that \abbrvSysOne can later use readily with little effort.
We note that \abbrvSysOne and \abbrvSysTwo are not systems in the multi-agent sense, but rather they encapsulate two wide classes of information processing.

\subsection{Classical and Multi-agent Epistemic Planning}

The idea of \emph{automated planning} has been present since the birth of AI and it has been widely explored in the computer science community ever since.
This area of research is a branch of artificial intelligence where the objective is to find plans, that is, sequences of actions, that can lead some acting agent(s) to achieve desired goals.

Its ``basic'' form is referred to as \emph{classical planning}.
In this setting, in order to have tractable and approachable problems, domains consider constrained environments, \ie they have to be
\begin{enumerate*}[label=\emph{\roman*)}]
    \item static;
    \item deterministic; and
    \item fully observable~\citep{ghallab2004automated}.
\end{enumerate*}
While is not our intention to provide a detailed explanation of the broad field of classical planning, we address the interested readers to~\citet{ghallab2004automated,modernApproach,bolander2011epistemic} for a complete introduction on the topic.

Epistemic planning, on the other hand, is a specific form of planning where the agents must additionally deal with the epistemic notions of knowledge and beliefs.
In the Multi-agent Epistemic Planning (MEP) problem, there are at least two planning agents. 
In what follows, we will only give an intuitive overview of MEP, referring the readers to~\citet{fagin1994reasoning,baral2021action} for details.
In this setting, we are concerned with finding the best series of actions that modifies the information flows to enable the agents to reach goals that refer to physical objects or agents' knowledge/beliefs \cite{bolander2011epistemic}.
We will use the term ``\knobel'' to encapsulate \textit{both} the notions of an agent's knowledge and their beliefs.
These concepts are  distinct in \emph{Dynamic Epistemic Logic} (DEL), which  is the underlying basis of the MEP problem, but for simplicity, in our high-level introduction, we will treat them as one.
%
In an MEP problem, we are concerned with the \knobel of the agents about the world or about others' \knobel.
This information is expressed through \emph{belief formulae}, \ie formulae that can express:
\begin{enumerate*}[label=\textit{\roman*})]
    \item physical properties of the worlds;
    \item \knobel of some agent (or group of agents) about these properties; and
    \item nested \knobel about others' \knobel.
\end{enumerate*}

The semantics of DEL formulae is traditionally expressed using \emph{pointed Kripke structures}~\cite{Kripke1963-KRISCO}.
The epistemic action language that we used in our work implements three \emph{types of action} and three \emph{observability relations} following the standard proposed by~\citet{baral2021action}.
The details of the language are not relevant to present our contribution, so to avoid unnecessary clutter, we will not present them.
Let us just note that each type of action defines a diverse transition function and alters an epistemic state in different ways.

Finally, the amount of information carried within a single epistemic state (\ie a \emph{Kripke structure}) and the high degree of liberty derived by complex transition functions (\eg~\citet{baral2021action,icaps20,DBLP:conf/pricai/FabianoBDPS21}) make planning in the multi-agent epistemic planning a very heavy-resource process that often brings to the unfeasibility of planning itself~\cite{bolander2015complexity}.
This is in contrast with classical planning where less intricate transition functions and state representation are used.

\section{Thinking Fast and Slow in Planning}
Two of the prominent lines of work in AI, \ie data-driven approaches and symbolic reasoning, seem to embody (even if loosely) the two \mysysS presented above.
In particular, data-driven approaches shares with \abbrvSysOne  the ability to build (possibly imprecise and biased) models from past experience, often represented by sets of data.
For example, perception activities, such as seeing, that in humans are handled by \abbrvSysOne, are currently addressed with machine learning techniques in AI.
Similarly, \abbrvSysTwo's capability to solve complex problems using a knowledge-based approach is somewhat emulated by AI techniques based on logic, search, and planning, which make use of explicit and well-structured knowledge.
While the parallelism data-driven--\abbrvSysOne and symbolic
--\abbrvSysTwo represent are a starting point in developing an automated fast and slow AI, we should not assume these two techniques to be the exclusive representative of the respective \mysys. 

In this paper, we transpose the concepts derived from the thinking fast and slow paradigm into the classical and multi-agent epistemic planning settings.
We will start by presenting general definitions for \abbrvSysOne and \abbrvSysTwo solvers and then describe the actual implementations of \abbrvSysOne and \abbrvSysTwo reasoners in these settings.
We will make use of three \emph{models} to represent key modules of our architecture, which from now on we will call \reasoner.
In particular, the \emph{model of self} is used to store the experience of the architecture, the \emph{model of the world} contains the knowledge accumulated by the system over the external environment and the expected tasks, while the \emph{model of others} contains the knowledge and beliefs about other agents who may act in the same environment.
Finally, the \emph{model updater} acts in the background to keep all models updated as new knowledge of the world, of other agents, or new decisions are generated and evaluated.

The general characterization of a \abbrvSysOne solver, triggered immediately when the problem is presented to \reasoner, does not require many factors.
These solvers are assumed to rely on the past experience of \reasoner itself.
Moreover, we assume that the running time for \abbrvSysOne approaches to be independent of the input and, instead, to depend on the experience accumulated by the overall architecture, in the \emph{model of self}.
Finally, we consider a \abbrvSysOne solver to be an entity that relies on \myvirgolette{intuition} (with a slight abuse of notation).
%
Taking into account these characteristics, the next question that naturally arises is \textit{can classical and epistemic planning ever be considered as \abbrvSysOne tasks, considering that planners, traditionally, always rely on look-ahead strategies?}
We considered some ideas that could help us develop a \abbrvSysOne planner.
Among those, only a few were not using search methods (intensively) but rather mostly relied on experience.  
Finally, we identified two feasible, yet functional, ways to exploit experience in the aforementioned planning settings.

The first approach makes use of \textit{pre-computed plans}; that is, \abbrvSysOne can be used to determine which of the plans already generated by past experiences is the one that \myvirgolette{fits the best} the current problem.
Of course, determining if an already computed plan is a good choice or not for the current problem is a difficult research question on its own.
Since the focus of this work is to devise a fast and slow architecture for planning rather than optimizing its internal components, we decided to use a simple, yet effective, criterion to select the best-fitting plan.
In particular, the first variation of \abbrvSysOne selects, among past solutions for the same domain, the pre-computed plan that is the closest in terms of \emph{Levenshtein Distance} and \emph{Jaccard Similarity}~\cite{8692996}, as we will see in more detail later.

Instead, the latter version of \abbrvSysOne, referred to as Plansformer~\cite{anonymous2023plansformer}, is based on a learning mechanism using LLMs pre-trained in coding languages such as Python, Java, and Ruby. The intuition behind selecting a code-based LLM is to inherit the syntactical knowledge, similar to the family of languages used to define a planning problem. Plansformer is fine-tuned on CodeT5 \citep{wang2021codet5} using planning problem instances and their corresponding plans. Given a new problem instance, Plansformer is capable of generating valid plans $\sim$90\% of the time, along with the confidence score for the generated plan. The confidence score is computed using the average non-zero probabilities of the generated tokens present in the plan given as output by Plansformer.  

Our \reasoner is a \abbrvSysOne-by-default architecture: whenever a new problem is presented, a \abbrvSysOne solver with the necessary skills to solve the problem starts working on it, generating a solution and a confidence level.
This allows to minimize the resource consumption making use of the much faster \abbrvSysOne solving process when there is no need for \abbrvSysTwo---that is when the solution proposed by \abbrvSysOne is \myvirgolette{good enough}.
Nevertheless, as for the human brain, \abbrvSysOne may encounter problems that it cannot solve, either due to its lack of experience or the inherent intricacy of the problem itself.
These situations require, then, the use of more thought-out resolution processes, generally provided by \abbrvSysTwo approaches.
Notice that we do not assume \abbrvSysTwo solvers to be always better than \abbrvSysOne solvers: given enough experience, some tasks could be better solved by \abbrvSysOne solvers. This behavior also happens in human reasoning~\citep{gigerenzer2009}.
Similarly, we don't assume \abbrvSysOne to be always more efficient than \abbrvSysTwo; in fact, if the problem is very simple it is possible that the ``well-thought'' process of \abbrvSysTwo could take less time than \abbrvSysOne.

As for \abbrvSysTwo we considered solving procedures, different depending on the setting, that employ traditional planning strategies.
For classical planning, we used the state-of-the-art \emph{Fast Downward} solver~\cite{helmert2006fast}.
In the case of \mep, instead, we decided to employ \EFP by~\citet{icaps20}.

\section{The Fast and Slow Paradigm}

\subsection{The SOFAI Architecture}
As the main contribution of this paper, we present an architecture\footnote{Link to repository removed to preserve the authors' anonymity.} inspired by cognitive theories to solve the classical and multi-agent epistemic planning problems.
In particular, our tool is heavily based on a recent architecture called SOFAI~\cite{sofai2021} 
that is, in turn, inspired by the dual-system proposed by~\citet{kahneman2011thinking}.
Following the ideas of Kahneman, the architecture is equipped with two types of \mysysS dedicated to computing a solution to an incoming task, and a third agent in charge of orchestrating the overall reasoning task.

In this architecture, incoming problems are initially handled by \abbrvSysOne solvers that have the required skills to tackle them.
\abbrvSysOne solvers compute a solution relying on the past experience collected by the architecture.
The computation is not affected by the size of the input problem and thus \abbrvSysOne solvers provide a solution in constant time.
Past experience is maintained in the model of the world, while the model of others maintains knowledge and beliefs over other agents that may act in the same environment.
A model updater agent is in charge of keeping all models updated as something new happens (\eg new decisions are generated).

The solution computed by the \abbrvSysOne solver (from now on for the sake of simplicity, let us assume it is just one \abbrvSysOne solver) and the corresponding confidence level is made available to the metacognitive (\MC) module which can now choose between the proposed solution or engaging a \abbrvSysTwo solver. An \abbrvSysTwo agent is typically a reasoning model that is able to deal with the current problem, this kind of solver is more demanding in terms of time and other types of resources. For this reason, only \MC\ agent can decide to activate an \abbrvSysTwo solver.

\MC\ agent assessment is based on several checks that are devoted to establishing whether it is worth adopting the solution proposed by \abbrvSysOne or engaging \abbrvSysTwo for a more accurate solution. This allows for minimizing time to action when there is no need for \abbrvSysTwo processing. 


\subsection{The Metacognitive Module}

For our MEP solver, following the SOFAI architecture~\citet{sofai2021}, we also defined a   \emph{metacognition process}.
This means that we want our \reasoner to be equipped with a set of mechanisms that allows it to both monitor and control its own cognitive activities, processes, and structures.
The goal of this form of control is to improve the quality of the system’s decisions. 
Metacognition models have been largely studied~\citep{Cox2005,Kralik2018,Kotseruba2020,Posner2020} in the past years.
Among the various proposed modalities, we envisioned our \reasoner to have a centralized metacognitive module that exploits both internal and external data and arbitrates between \abbrvSysOne and \abbrvSysTwo solvers.
Let us note that this module is structurally and conceptually different from an algorithm portfolio selection~\citep{DBLP:journals/ec/KerschkeHNT19}.

We propose a metacognitive module that itself follows the \emph{thinking fast and slow} paradigm.
This means that our \MC\ module is comprised of two main phases: the first one takes intuitive decisions without considering many factors, while the second one is in charge of carefully selecting the best-solving strategy, considering all the available elements whenever the first phase did not manage to return an adequate solution.
We will refer to the former with \MCone and to the latter with \MCtwo.

\MCone  is in charge of deciding whether to accept  the solution proposed by the \abbrvSysOne solver or to activate \MCtwo.
\MCone takes this decision considering the \emph{confidence}, among other few factors, of the \abbrvSysOne solver: if the confidence, which usually depends on the amount of experience, is high enough, \MCone adopts the \abbrvSysOne solver’s solution. 

If \MCone decides that the solution of the \abbrvSysOne solver is not \myvirgolette{good enough}, it engages \MCtwo.
Intuitively, this module needs to evaluate whether to accept the solution proposed by the \abbrvSysOne solver or which \abbrvSysTwo solver to activate for the task. 
To do this, \MCtwo compares the expected reward for the \abbrvSysTwo solver with the expected reward of the \abbrvSysOne one: if the expected additional reward of running the \abbrvSysTwo solver, compared to the \abbrvSysOne one, is large enough, then \MCtwo activates the \abbrvSysTwo solver.
\MCtwo, following the human reasoning model~\citep{Shenhav2013}, is designed to avoid costly reasoning processes unless the additional cost is compensated by an even greater expected reward for the solution that the \abbrvSysTwo solver will devise.

\section{Metacognition at Work}

In what follows, we provide a \myvirgolette{concrete} view of the \abbrvSysOne/\abbrvSysTwo framework for the above-mentioned settings.
To do so, we will make use of Algorithms~\ref{alg:metacog}--\ref{alg:exec_stwo}.

Before going into detail, let us briefly comment on the input and on the parameters of these algorithms.
The process requires the domain description (\varstyle{D}), a particular instance (\varstyle{I}) that we want to solve on such domain, and the time limit (\varstyle{TL}) within which the instance needs to be solved.
The parameters, instead, represent some internal values that capture some sort of \myvirgolette{inclination} of the architecture towards employing \abbrvSysOne.
In particular, we have that:
\begin{enumerate*}[label=\textit{\roman*)}]
\item the acceptable correctness (\varstyle{A}) represents the minimum ratio of solved goals, w.r.t. the total number of them, that defines an acceptable solution. Let us note that this measure can also be changed to depend on other factors or to account for goals' importance, for example. Its default value is 0.5;
\item \varstyle{T1} represents the minimal amount of experience required by \reasoner to consider a solution proposed by \abbrvSysOne. Its default value is set to 20;
\item \varstyle{T2} represents the minimum number of \abbrvSysOne usages after which it will consider \abbrvSysOne accountable for its mistakes. This threshold allows the architecture to initially try to employ \abbrvSysOne more freely, to augment its experience.
Conversely, after the minimum number \varstyle{T2} of solutions, the metacognition actually uses the previous performances of \abbrvSysOne to check for \abbrvSysOne accountability. Its default value is set to 20;
\item \varstyle{T3} is a value between 0 and 1 and it is used to represent the \emph{risk-aversion} of the architecture: the higher the value the more incline \reasoner is to use \abbrvSysTwo. The default value is set to 0.6;
\item $\epsilon$ is a factor that is used to scale the probability that \abbrvSysOne solution may actually be employed even if it was not considered convenient.
This is added to increase the number of \abbrvSysOne usages, and consequently its experience, in those situations where the low confidence of \abbrvSysOne itself may limit it too aggressively.
Let us note that the solution proposed by \abbrvSysOne needs to be validated before being accepted in any case (Line 2 of Algorithm~\ref{alg:exec_sone}).
Its default value is 0.1;
\item (\varstyle{M}) represents the experience of \reasoner. Every time a solution for a problem is found, this is stored in the memory with a series of useful information, \eg the correctness value, the employed system (\ie \abbrvSysOne or \abbrvSysTwo), the difficulty of the instance, the required time, and so on.
\end{enumerate*}

\begin{algorithm}[!htb]
\caption{Fast and Slow MEP Architecture}
\label{alg:metacog}
\textbf{Input}: Domain (\varstyle{D}), Instance (\varstyle{I}), Time Limit (\varstyle{TL})\\
\textbf{Parameter}: Acceptable Correctness (\varstyle{A}), \varstyle{T1}, \varstyle{T2}, \varstyle{T3}, $\epsilon$, Memory (\varstyle{M}),\\
\textbf{Output}: Plan (\varstyle{S}), Correctness (\varstyle{C})
\begin{algorithmic}[1] 
\State Let \varstyle{p} be the solution of \varstyle{I} found by \abbrvSysOne
\State Let \varstyle{cx} be the confidence of \abbrvSysOne on  \varstyle{p}.%
\If {$|$\varstyle{M}.\funcstyle{solved\_instances(\varstyle{D})}$|$ $\geq$ \varstyle{T1}}
\If {$|$\varstyle{M}.\funcstyle{solved\_instances(\varstyle{D},\abbrvSysOne)}$|$ $<$ \varstyle{T2}}
\State Let \varstyle{K} $ = 0$
\Else
\State Let \varstyle{avg\_corr} $= 0$
\ForAll{\varstyle{i} $\in$ \varstyle{M}.\funcstyle{solved\_instances(\varstyle{D},\abbrvSysOne)}}
\State \varstyle{avg\_corr} $+= \frac{|\varstyle{i}.\funcstyle{solved\_goals()}|}{|\varstyle{i}.\funcstyle{tot\_goals()}|}$
\EndFor
\State Let \varstyle{K} $= 1-$ \varstyle{avg\_corr}
\EndIf
\If{\varstyle{cx} $\times $ $(1-\varstyle{K})$ $\geq$ \varstyle{T3}}
\State \Return{$\langle$\varstyle{S},\varstyle{C}$\rangle$ $=$ \funcstyle{try\_S1(\varstyle{p},\varstyle{D},\varstyle{I},\varstyle{TL})}}
\EndIf
\EndIf
\State Let \varstyle{diff} $=$ \varstyle{I}.\funcstyle{compute\_difficulty()}
\State Let \varstyle{est\_t} $=$ \varstyle{M}.\funcstyle{avg\_t\_from\_diff(\varstyle{diff})} 
\State Let \varstyle{rem\_t} $=$ \varstyle{TL} $-$ elapsed\_t
\State Let \varstyle{est\_cost} $=$ $\frac{\varstyle{est\_t}}{\varstyle{rem\_t}}$
\If{\varstyle{est\_cost} $> 1$}
\State \Return{$\langle$\varstyle{S},\varstyle{C}$\rangle$ $=$ \funcstyle{try\_S1(\varstyle{p},\varstyle{D},\varstyle{I},\varstyle{TL})}}
\Else
\State Let \varstyle{prob} $=$ $(1-\varstyle{T3}) \times \epsilon$
\If{\varstyle{prob} $\geq$ \funcstyle{generate\_random\_number(0,1)}}
\State \Return{$\langle$\varstyle{S},\varstyle{C}$\rangle$ $=$ \funcstyle{try\_S1(\varstyle{p},\varstyle{D},\varstyle{I},\varstyle{TL})}}
\Else
\State \varstyle{C} $= \frac{|\varstyle{I}.\funcstyle{solved\_goals(p)}|}{|\varstyle{I}.\funcstyle{tot\_goals()}|}$
\If{\varstyle{C} $\geq$ \varstyle{A}}
\If{$(1-(\varstyle{est\_cost} \times (1-\varstyle{T3}))$ $\geq$ $\varstyle{C}\times (1-\varstyle{K})$}
\State \Return{$\langle$\varstyle{S}, \varstyle{C}$\rangle$ $=$ solve\_with\_S2(\varstyle{p},\varstyle{D},\varstyle{I},\varstyle{rem\_t})}
\Else
\State \Return{$\langle$\varstyle{S} $=$ \varstyle{p},\varstyle{C}$\rangle$}
\EndIf
\Else
\State \Return{$\langle$\varstyle{S}, \varstyle{C}$\rangle$ $=$ solve\_with\_S2(\varstyle{null},\varstyle{D},\varstyle{I},\varstyle{rem\_t})}
\EndIf
\EndIf
\EndIf
\end{algorithmic}
\end{algorithm}

We are now ready to describe in more detail Algorithms~\ref{alg:metacog}--\ref{alg:exec_stwo}.
Let us start by presenting Algorithms~\ref{alg:metacog}.
In particular, we can identify \MCone in Lines 1--16, and \MCtwo in Lines 17--39.
As already mentioned, to better emulate the thinking fast and slow paradigm, we assume \sysone to automatically start and provide a solution at the beginning of \MCone.
That is why we start the metacognitive process by storing the results of such process in the variables \varstyle{p} and \varstyle{cx}, which represent the solution found by \abbrvSysOne and the confidence that \abbrvSysOne has about this solution appropriateness, respectively.
The metacognitive process then proceeds to check whether the experience accumulated by the architecture is enough to consider \abbrvSysOne reliable (Line 3).
If the architecture has enough experience, the metacognition considers the confidence of \abbrvSysOne, adjusted to take into account the previous solutions proposed by \abbrvSysOne itself (Lines 4--12), and determines whether \abbrvSysOne's confidence is within the tolerated risk, identified by \varstyle{T3} (Line 13).
If the confidence of \abbrvSysOne is enough, then \reasoner tries to employ \abbrvSysOne's solution (line 14).

If at any point, \abbrvSysOne's solution is considered not appropriate by the metacognitive process---because it violates some checks---then \MCtwo starts.
This part of the procedure begins by determining a value that represents the difficulty of the problem instance (derived by various factors such as the number of agents, possible actions, fluents, and so on) at Line 17.
This measure is then used to determine the average solving time for a given difficulty and to estimate the cost of solving the given problem (Line 18--20).
If the cost exceeds $1$ then there is not enough time to call \abbrvSysTwo and \reasoner tries to employ \abbrvSysOne.
The system can also adopt \abbrvSysOne with a probability that is related to the risk aversion \varstyle{T3} and a parameter $\epsilon$, this is done in Lines 24--26, to improve the exploration skill of the architecture itself.
\reasoner evaluates the solution proposed by \abbrvSysOne and, if it is acceptable (Line 29),  whether the extra time required by \abbrvSysTwo counterbalanced the cost (Line 30).
If the solution is not acceptable or the increase in correctness is big enough \abbrvSysTwo is called (Line 36 and 31, respectively), otherwise the solution proposed by \abbrvSysOne is used (Line 33).

\begin{algorithm}[!htb]
\caption{\funcstyle{try\_S1} function}
\label{alg:exec_sone}
\textbf{Input}: Plan (\varstyle{p}), Domain (\varstyle{D}), Instance (\varstyle{I}), Time Limit (\varstyle{TL})\\
\textbf{Parameter}: Acceptable Correctness (\varstyle{A})\\
\textbf{Output}: Plan (\varstyle{S}), Correctness (\varstyle{C})
\begin{algorithmic}[1] 
\State \varstyle{C} $= \frac{|\varstyle{I}.\funcstyle{solved\_goals(p)}|}{|\varstyle{I}.\funcstyle{tot\_goals()}|}$
\If{\varstyle{C} $\geq$ \varstyle{A}}
\State \Return{$\langle$\varstyle{S} $=$ \varstyle{p}, \varstyle{C}$\rangle$}
\Else 
\State \Return{$\langle$\varstyle{S}, \varstyle{C}$\rangle$ $=$ solve\_with\_S2(\varstyle{null},\varstyle{D},\varstyle{I},\varstyle{TL})}
\EndIf
\end{algorithmic}
\end{algorithm}

\begin{algorithm}[!htb]
\caption{\funcstyle{solve\_with\_S2} function}
\label{alg:exec_stwo}
\textbf{Input}: Plan (\varstyle{p}), Domain (\varstyle{D}), Instance (\varstyle{I}) Time Limit (\varstyle{TL})\\
\textbf{Output}: Plan (\varstyle{S})
\begin{algorithmic}[1] 
\If{\abbrvSysTwo(\varstyle{D},\varstyle{I}) terminates within \varstyle{TL}}
\State \Return{$\langle$\varstyle{S} $=$ \abbrvSysTwo.get\_plan(), \varstyle{C} $=1\rangle$}
\ElsIf{\varstyle{p} $!=$ \varstyle{null}}
\State \Return{$\langle$\varstyle{S} $=$ \varstyle{p}, \varstyle{C} $=\frac{|\varstyle{I}.\funcstyle{solved\_goals(p)}|}{|\varstyle{I}.\funcstyle{tot\_goals()}|}\rangle$}
\Else
\State \textbf{OPT-OUT}
\EndIf
\end{algorithmic}
\end{algorithm}

Algorithms~\ref{alg:exec_sone} and~\ref{alg:exec_stwo} are instead used to try and adopt the solution proposed by \abbrvSysOne and to try and solve the problem with \abbrvSysTwo, respectively.
In particular, Algorithms~\ref{alg:exec_sone} takes the solution proposed by \abbrvSysOne and checks whether it has an acceptable degree of correctness. If it does then the solution is employed, otherwise, Algorithm~\ref{alg:exec_stwo} is called.
This function simply calls the \abbrvSysTwo approach (\ie Fast Downward or \EFP, depending on the setting) on the instance of the problem to solve and, if it terminates before the available time ends, it returns the plan found by the \abbrvSysTwo planner with confidence equal to 1. If \abbrvSysTwo cannot find the solution within the time limit then the solution from \abbrvSysOne, if acceptable, is adopted; otherwise, \reasoner returns no solution and terminates.

\subsection{\abbrvSysOne and \abbrvSysTwo solvers}
While in the previous paragraph, we described how our architecture decides which solving approach is the most appropriate, here we will provide a high-level overview of solving processes themselves. 
In particular, we designed our \abbrvSysOne solvers to solely rely on past experience.

The first type of \abbrvSysOne solver, which is the case-based one, analyzes the memory and looks, through the various solved instances (using either \abbrvSysOne itself or \abbrvSysTwo), which one is the closest to the problem that is being tackled.
Once the closest instance is identified, \abbrvSysOne returns the plan associated with it as a solution and the distance value as a measure for confidence.
This distance can be calculated in two different ways, generating effectively two different \abbrvSysOne solvers\footnote{We compare the performances of these two \abbrvSysOne solvers later in the paper.}.
The first measure of distance considers the problems as a set of formulae, \ie the ones that comprise the initial and goal states, and adopts the well-known Jaccard Similarity~\cite{8692996}, that is the ratio between the union and the intersection of the two sets we are considering, as a metric for finding similarity between the input problem and the instances existing in the case library.
The second metric is calculated by transforming the two instances into two distinct strings, once again comprised of all the initial and goal states information (separated by the special characters \myvirgolette{$|$}), that are then compared using the Levenshtein distance~\cite{haldar2011levenshtein} to determine the actual distance measure.
Let us note that since we are considering only instances of the same domain, there is no need to incorporate other information.
Nonetheless, if we would like to compare also instances of different domains, the domains' descriptions could easily be added to the representative of each instance.

The other type of \abbrvSysOne takes the domain and problem description of a planning instance and maps it to a sequential prompt given as input to Plansformer~\cite{anonymous2023plansformer}. The output obtained is a plan along with the associated confidence score.
Let us note that this \abbrvSysOne makes use of a pre-computed training set and does not increase its experience during the solving phase.
This is a design choice and we leave the exploration of Plansformer with ``dynamic" training as a future work.
Furthermore, Plansformer currently can only handle classical planning problems and not epistemic ones.
This is because of Plansformer's limitations on the input token length, which is currently 512.
In fact, MEP problems given their higher intricacy, when converted to the prompt desired by Plansformer, cross the 512 tokens limit.
Once again, we working to solve this issue is but leave the full integration of Plansformer in MEP as future work.

Finally, while for \abbrvSysOne we needed to implement ad-hoc solutions, the same is not true for \abbrvSysTwo planners.
In fact, as mentioned we employed the state-of-the-art classical and epistemic planners, \ie Fast Downward and \EFP, as our \abbrvSysTwo solvers.
Given that explaining how these planners work is beyond the scope of this paper, we can safely assume these approaches to be black boxes that return the best possible solutions, if exist.
Nonetheless, we refer the interested readers to~\citet{helmert2006fast,icaps20} for a detailed explanation of the internal mechanisms of Fast Downward and \EFP, respectively.

\section{Experimental Results and Discussion}

\subsection{Experimental Setup}
In this section, we compare the new architecture introduced as the main contribution of this paper with Fast Downward~\cite{helmert2006fast} and \EFP~\cite{icaps20} that, to the best of our knowledge, are the state-of-the-art solver in the respective fields.
All the experiments were performed on a 3.00GHz Intel Core i7-5500U machine with 16GB of memory.

As benchmarks, we used two distinct domains, depending on the type of planning we wanted to analyze.
To evaluate classical planning we employed the well-known \textbf{Blocks-World} (\textbf{BW}) domain \citep{gupta1991complexity}.
In this domain, the acting agent is a \emph{mechanical arm} that can move \emph{blocks} and can determine whether it is holding one or not.
The mechanical arm can only hold, and therefore move, one block at the same time; and a block can only be placed on top of a \emph{clear} block---a block with no blocks on top of it and that is not held by the mechanical arm---or on the table.
The goals in this domain refer to specific configurations of the blocks.

For MEP we employed a small variation of the standard epistemic planning domain known as \textbf{Coin in the Box} (\textbf{CB})~\cite{kominis2015beliefs,huang2017general,baral2021action}.
In this domain, $n \geq 2$ agents are in one of two rooms, one of which contains a box with a coin inside.
In the initial configuration, 
everybody knows that:
\begin{enumerate*}[label=\textit{\roman*)}]
\item none of the agents know whether the coin lies heads or tails up;
\item the box is locked;
\item only one agent has the key that opens the box;
\item each agent might be attentive or not w.r.t. to certain actions execution;
\end{enumerate*}
Moreover, we know that each agent can execute one of the following actions:
\begin{enumerate*}[label=\textit{\arabic*)}]
\item \texttt{move}: an agent can move to the other room;
\item \texttt{open}: an agent, if it has the key, can open the box;
\item \texttt{peek}: to learn whether the coin lies heads or tails up, an agent can peek into the box, but this requires the box to be open;
\item \texttt{announce}: this will result in all the listening (\ie attentive) agents to
believe that the coin lies heads or tails up depending on the announced
value;
\item \texttt{distract/signal} another agent: these actions will make an attentive agent no more attentive or vice-versa, respectively.
\end{enumerate*}
The goals usually consist of some agents knowing whether the coin lies heads or tails up while other agents know that it knows, or are ignorant about this.

The experiments for classical planning are comprised of 700 
different problems that vary the initial state, goal state, and a number of blocks.
Of these, 150 are instances with 4 blocks, 150 with 5, 100 with 10, 100 with 11, 100 with 12, and 100 with 13. 
Similarly, for epistemic planning, we used a set of 240 different instances of the \textbf{CB} domain\footnote{Let us note that given the intricacy of epistemic reasoning, we opted for fewer instances to have reasonable testing times.}.

Regarding the various input and parameters of the architecture (used in Algorithms~\ref{alg:metacog},~\ref{alg:exec_sone},~\ref{alg:exec_stwo}) we imposed:
\begin{enumerate*}[label=\emph{\roman*)}]
\item a Time Limit (\varstyle{TL}) of 60 and 90 seconds to solve each instance for classical and epistemic, respectively;
\item an Acceptable Correctness (\varstyle{A}) of 0.5, meaning that at least half of the goals must be satisfied for a \abbrvSysOne solution to be considered;
\item the various thresholds (\ie \varstyle{T1}, \varstyle{T2}, \varstyle{T3}) and  $\epsilon$ to have their default values; and
\item the Memory (\varstyle{M}) to be initially filled with 25 already solved problems in the case of classical planning and empty for MEP.
\end{enumerate*}

\subsection{Results}

As a baseline for our experiments, we used the solvers Fast Downward~\cite{helmert2006fast} and  \EFP~\cite{icaps20}.
As mentioned, we let the solvers tackle all the instances with a time-out of 60/90 seconds per instance.
The main idea is that these approaches represent the current capabilities of classical and epistemic planning and are comparable to a solely \abbrvSysTwo-based architecture.
While these approaches are guaranteed to find the best solution, if this exists, they are not flexible enough to adapt to situations where the resources, namely time, are limited.

\begin{figure*}[ht]
	\centering
	\includegraphics[width=0.85\linewidth]{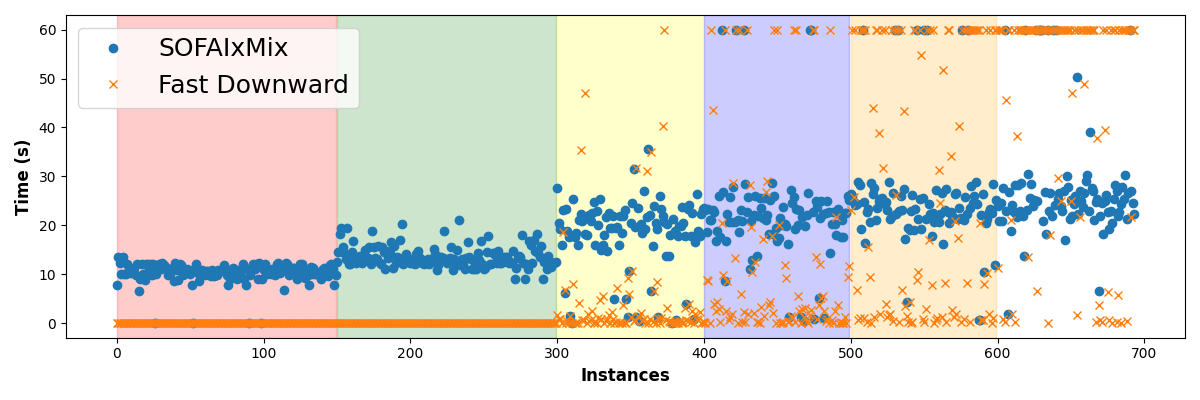}
	\caption{Time comparison between \texttt{Mix} and Fast Downward.}
	\label{fig:time_comp_cls}
	%
\end{figure*}

\begin{figure*}[ht]
	\centering
	\includegraphics[width=0.85\linewidth]{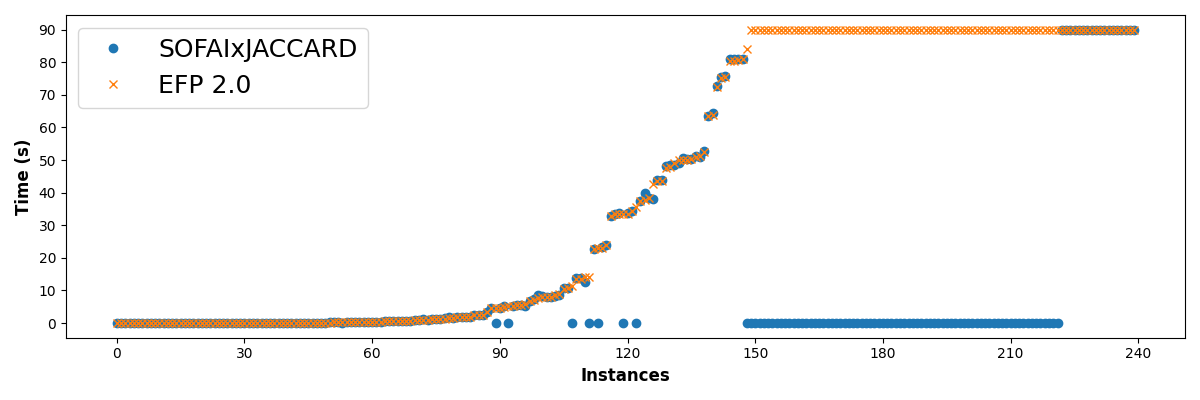}
	\caption{Time comparison between \texttt{Jac} and \EFP.}
	\label{fig:time_comp}
	%
\end{figure*}

We then compared the baseline results with different configurations of the architecture presented in this paper (Tables~\ref{tab:res_cls} and \ref{tab:res}).
To avoid unnecessary clutter, let us identify the various configurations, with the following abbreviations:
\begin{itemize}
    \item \texttt{Jac}: the configuration of the architecture where the \abbrvSysOne makes use of the Jaccard similarity as a notion of distance;
    \item \texttt{Lev}: the configuration where the metric for the distance between problems is defined through Levenshtein distance (not reported in classical as it performed worse than \texttt{Jac} in every aspect);
    \item \texttt{Plf}: the configuration where \abbrvSysOne employs Plansformer as \abbrvSysOne.
    \item \texttt{Mix}: the configuration where \abbrvSysOne chooses the most similar instance in the memory, w.r.t. the given problem, by selecting the highest (normalized) score between the ones calculated with all the available \abbrvSysOne in the architecture (let us recall that we could not employ Plansformer to tackle MEP problems);
    \item \texttt{Rng}: the configuration where the \abbrvSysOne randomly picks one of the instances in memory as \myvirgolette{most similar} without even considering a notion of distance. This approach was inserted as a baseline to outperform with a well-thought \abbrvSysOne.
\end{itemize}



\begin{table}[!htb]
    \centering
	\resizebox{1.0\columnwidth}{!}{%
	    \begin{tabular}{||l|c|c|c|c|c||}
		    \hhline{|t:======:t|}
			& \footnotesize{\textbf{FD}}  & \texttt{Jac}   & \texttt{Plf}  & \texttt{Mix} & \texttt{Rng}                                \\
			\hhline{||------||}
			Solved   & 586 & 593     & \textbf{671}    & \textbf{671}   & 587       \\
			 & & (+5\%) & (+19\%)
			 & (+19\%) & (+0.2\%)\\ 
			\hhline{||------||}
			Time (avg)  & 12.289  & \textbf{11.609} & 18.687 & 18.480 & 12.232 \\
  			\hhline{||------||}
			Corr (avg)   & 0.84       & 0.79 & 0.89 & \textbf{0.90} &   0.80       \\
			\hhline{||------||}
            \abbrvSysOne calls   & 0       & 151 & 626 & 624 &  68                    \\
			\hhline{|b:======:b|}
		\end{tabular}
	}
	\caption{Comparison between Fast Downward (\textbf{FD}) and the diverse configurations of \reasoner on various parameters. The row Solved expresses the number of instances, out of the 700, that have been solved with correctness at least superior to 0.5. Time is expressed in seconds and the correctness (Corr), being the ratio between solved goals and their total number, is a measure between 0 and 1. All the results also consider the instances that could not be solved, to these we assigned 60 seconds as time and 0 as correctness.}
	\label{tab:res_cls}%
\end{table}

\begin{table}[!htb]
    \centering
	\resizebox{1.0\columnwidth}{!}{%
	    \begin{tabular}{||l|c|c|c|c|c||}
		    \hhline{|t:======:t|}
			& \EFP  & \texttt{Jac}   & \texttt{Lev}  & \texttt{Mix} & \texttt{Rng}                                \\
			\hhline{||------||}
			Solved   & 149 & \textbf{222}      & 207    & 197   &181       \\
			 & & (+49\%) & (+39\%)
			 & (+32\%) & (+21\%)\\ 
			\hhline{||------||}
			Time (avg)  & 42.737  & \textbf{14.495} & 19.608 & 20.001 & 27.755 \\
  			\hhline{||------||}
			Corr (avg)   & 0.62       & \textbf{0.81} & 0.89 & 0.67 &   0.66       \\
			\hhline{||------||}
            \abbrvSysOne calls   & 0       & 86 & 74 & 115 &  69                     \\
			\hhline{|b:======:b|}
		\end{tabular}
	}
	\caption{Comparison between \EFP\ and the diverse configurations of our architecture on various parameters. The table schema follows Table~\ref{tab:res_cls}. In this setting, we assigned 90 seconds to the unsolved instances.}
	\label{tab:res}%
\end{table}

Tables~\ref{tab:res_cls} and \ref{tab:res} show that \texttt{Mix} and \texttt{Jac} (for classical and epistemic, respectively) are the best all-around configurations, with the most number of solved instances, the lowest average solving times and the highest average correctness of \reasoner. 
To provide further analysis we present, in Figures~\ref{fig:time_comp_cls} and \ref{fig:time_comp}, the plots that show the resulting times of these configurations against Fast Downward and \EFP.
In these, the instances (x-axis) are sorted w.r.t. the notion of ``difficulty" and, in the particular case of classical planning we grouped the instances with the same number of blocks\footnote{4 in pink, 5 in green, 10 in yellow, 11 in purple, 12 in orange, and 13 in white.}.
The results for classical planning show that when the problem is ``simple enough" Fast Downward outperforms \reasoner but, as soon as the instances grow in difficulty, the constant time required by \abbrvSysOne begins to outperform Fast Downward making \reasoner a reliable tool to solve the hardest configurations in a time-constrained environment.
In the epistemic settings, instead, all the solving time of \texttt{Jac} outperforms \EFP (equal or lower height on the y-axis). 
Let us note that the instances which are placed exactly on the top of the plot, \ie on the 60/90 seconds line, are the ones that have timed out.

\subsection{Discussion}
Tables~\ref{tab:res_cls} and \ref{tab:res} and Figures~\ref{fig:time_comp_cls} and \ref{fig:time_comp} highlight some interesting results about our architecture.
In particular, 
thanks to its ability to \myvirgolette{adapt} to resources constraints, our architecture can provide a flexible tool to solve instances even for those settings where the inherent complexity of the problems (\eg multi-agent epistemic planning) brings unfeasibility to the solving process.
Our proposal can be seen as a way to exploit the accumulated experience, in a human fashion, to quickly solve known problems that otherwise would require a lot of effort.

Moreover, the trade-off offered by allowing not-fully correct, but sound, plans is also a way to produce a solution that, even if partial, can be more useful than no solution at all.
Let us note that some of the solved problems by \abbrvSysOne, and not by \abbrvSysTwo, do not have a solution that can satisfy all the goals.
While, for example in MEP, the planning process cannot detect these situations given the undecidability of MEP~\cite{bolander2015complexity}, our architecture simply reports a plan to reach some of the subgoals, providing once again a better alternative to no plan at all.
Nonetheless, the parametric nature of the architecture allows also us to tweak it so that it only accepts fully correct plans.
This can be done by setting the value of the acceptable correctness (\varstyle{A}) to 1.
With this small change, the architecture will only return plans that reach the goal state while still taking advantage of the \abbrvSysOne capabilities of the architecture (albeit \abbrvSysOne solutions would be adopted fewer times).
The same goes for the other internal parameters that easily allow the user to modify the architecture so that it is more prone to accept less accurate solutions in favor of saving time, and vice-versa.

Another interesting result that we can deduce from Tables~\ref{tab:res_cls} and \ref{tab:res} is that the employment of \texttt{Mix} is worse than both \texttt{Jac} and \texttt{Lev} in epistemic while it is the best configuration in classical.
While, at first glance, this might seem a contradictory result it actually shows that the balance between the usage of \abbrvSysOne and \abbrvSysTwo need to be preserved by the architecture.
In fact, as \texttt{Mix}, in epistemic, uses the combination of \texttt{Jac} and \texttt{Lev}, the number of times that a solution of \abbrvSysOne is employed is higher than the two approaches (115 against 86 and 74).
This makes it so that the architecture has fewer opportunities to increase its experience that can be re-used later to solve different problems.
Contrarily in classical, \texttt{Plf} does not rely on the knowledge accumulated during the solving phase and remains a viable tool during all the resolution process making the additions provided by \texttt{Jac} an enrichment to the overall performances.
While it is not easy to find the best trade-off for limiting the employment of \abbrvSysOne, it is our intention to define ways to automatically tune the internal parameters of the architecture so that it can exploit at maximum its capabilities.

\section{Conclusions and Ongoing Work}
In this work, we presented an architecture to tackle planning problems, in different settings, that is heavily inspired by the well-known cognitive theory Thinking Fast and Slow by~\citet{kahneman2011thinking}. This tool builds on the SOFAI architecture~\cite{sofai2021}, which makes use of a metacognitive process to arbitrate the solving processes, and two solvers referred to as \sysone and \systwo. While \abbrvSysTwo is directly derived from the literature the \abbrvSysOne solvers have been designed ad-hoc for the proposed architecture to exploit past experience.
The SOFAI-inspired approach showed very promising results outperforming the state-of-the-art planners in different metrics.
Another advantage of the proposed architecture is that it can be used to incorporate new solving techniques developed by the community. In fact, our tool can be easily modified to employ different \abbrvSysOne or \abbrvSysTwo, or multiple versions of them. We are currently working on a version of the architecture that allows for multiple \abbrvSysTwo in epistemic planning, in particular both \EFP and RP-MEP~\cite{muise2015planning}.

\bibliographystyle{unsrtnat}  
\bibliography{sofaiplanning.bib}

\begin{thebibliography}{31}
\providecommand{\natexlab}[1]{#1}
\providecommand{\url}[1]{\texttt{#1}}
\expandafter\ifx\csname urlstyle\endcsname\relax
  \providecommand{\doi}[1]{doi: #1}\else
  \providecommand{\doi}{doi: \begingroup \urlstyle{rm}\Url}\fi

\bibitem[Booch et~al.(2021)Booch, Fabiano, Horesh, Kate, Lenchner, Linck,
  Loreggia, Murugesan, Mattei, Rossi, and Srivastava]{booch-2021-thinkfast}
Grady Booch, Francesco Fabiano, Lior Horesh, Kiran Kate, Jonathan Lenchner,
  Nick Linck, Andrea Loreggia, Keerthiram Murugesan, Nicholas Mattei, Francesca
  Rossi, and Biplav Srivastava.
\newblock Thinking fast and slow in {AI}.
\newblock In \emph{Proceedings of the 35th AAAI conference}, pages
  15042--15046, 2021.

\bibitem[Kahneman(2011)]{kahneman2011thinking}
Daniel Kahneman.
\newblock \emph{Thinking, Fast and Slow}.
\newblock Macmillan, 2011.

\bibitem[Pallagani et~al.(2022)Pallagani, Muppasani, Murugesan, Rossi, Horesh,
  Srivastava, Fabiano, and Loreggia]{anonymous2023plansformer}
Vishal Pallagani, Bharath Muppasani, Keerthiram Murugesan, Francesca Rossi,
  Lior Horesh, Biplav Srivastava, Francesco Fabiano, and Andrea Loreggia.
\newblock Plansformer: Generating symbolic plans using transformers, 2022.
\newblock URL \url{https://arxiv.org/abs/2212.08681}.

\bibitem[Marcus(2020)]{marcus2020decade}
Gary Marcus.
\newblock The next decade in ai: Four steps towards robust artificial
  intelligence, 2020.

\bibitem[Newell(1992)]{newell1992soar}
Allen Newell.
\newblock {SOAR} as a unified theory of cognition: Issues and explanations.
\newblock \emph{Behavioral and Brain Sciences}, 15\penalty0 (3):\penalty0
  464--492, 1992.

\bibitem[Goel et~al.(2017)Goel, Chen, and Wierman]{goel2017thinking}
Gautam Goel, Niangjun Chen, and Adam Wierman.
\newblock Thinking fast and slow: Optimization decomposition across timescales.
\newblock In \emph{2017 IEEE 56th Annual Conference on Decision and Control
  (CDC)}, pages 1291--1298. IEEE, 2017.

\bibitem[Ganapini et~al.(2021)Ganapini, Campbell, Fabiano, Horesh, Lenchner,
  Loreggia, Mattei, Rossi, Srivastava, and Venable]{sofai2021}
Marianna~Bergamaschi Ganapini, Murray Campbell, Francesco Fabiano, Lior Horesh,
  Jon Lenchner, Andrea Loreggia, Nicholas Mattei, Francesca Rossi, Biplav
  Srivastava, and Kristen~Brent Venable.
\newblock Thinking fast and slow in {AI:} the role of metacognition.
\newblock \emph{CoRR}, abs/2110.01834, 2021.
\newblock URL \url{https://arxiv.org/abs/2110.01834}.

\bibitem[Kotseruba and Tsotsos(2020)]{Kotseruba2020}
Iuliia Kotseruba and John~K. Tsotsos.
\newblock 40 years of cognitive architectures: core cognitive abilities and
  practical applications.
\newblock \emph{Artificial Intelligence Review}, 53\penalty0 (1):\penalty0
  17--94, Jan 2020.
\newblock \doi{10.1007/s10462-018-9646-y}.

\bibitem[Ghallab et~al.(2004)Ghallab, Nau, and Traverso]{ghallab2004automated}
Malik Ghallab, Dana Nau, and Paolo Traverso.
\newblock \emph{Automated Planning: theory and practice}.
\newblock Elsevier, 2004.

\bibitem[Russell and Norvig(2010)]{modernApproach}
Stuart~J. Russell and Peter Norvig.
\newblock \emph{Artificial Intelligence - {A} Modern Approach, Third
  International Edition}.
\newblock Pearson Education, 2010.
\newblock ISBN 978-0-13-207148-2.
\newblock URL
  \url{http://vig.pearsoned.com/store/product/1,1207,store-12521\_isbn-0136042597,00.html}.

\bibitem[Bolander and Andersen(2011)]{bolander2011epistemic}
Thomas Bolander and Mikkel~Birkegaard Andersen.
\newblock Epistemic planning for single-and multi-agent systems.
\newblock \emph{Journal of Applied Non-Classical Logics}, 21\penalty0
  (1):\penalty0 9--34, 2011.
\newblock \doi{10.1016/0010-0277(83)90004-5}.

\bibitem[Fagin et~al.(1995)Fagin, Moses, Halpern, and
  Vardi]{fagin1994reasoning}
Ronald Fagin, Yoram Moses, Joseph~Y. Halpern, and Moshe~Y. Vardi.
\newblock \emph{Reasoning About Knowledge}.
\newblock MIT press, 1995.
\newblock ISBN 9780262061629.

\bibitem[Baral et~al.(2022)Baral, Gelfond, Pontelli, and Son]{baral2021action}
Chitta Baral, Gregory Gelfond, Enrico Pontelli, and Tran~Cao Son.
\newblock An action language for multi-agent domains.
\newblock \emph{Artificial Intelligence}, 302:\penalty0 103601, 2022.
\newblock ISSN 0004-3702.
\newblock \doi{https://doi.org/10.1016/j.artint.2021.103601}.
\newblock URL
  \url{https://www.sciencedirect.com/science/article/pii/S0004370221001521}.

\bibitem[Kripke(1963)]{Kripke1963-KRISCO}
Saul~A. Kripke.
\newblock Semantical analysis of modal logic i normal modal propositional
  calculi.
\newblock \emph{Mathematical Logic Quarterly}, 9\penalty0 (5-6):\penalty0
  67--96, 1963.
\newblock \doi{10.1002/malq.19630090502}.

\bibitem[Fabiano et~al.(2020)Fabiano, Burigana, Dovier, and Pontelli]{icaps20}
Francesco Fabiano, Alessandro Burigana, Agostino Dovier, and Enrico Pontelli.
\newblock {EFP} 2.0: {A} multi-agent epistemic solver with multiple e-state
  representations.
\newblock In \emph{Proceedings of the Thirtieth International Conference on
  Automated Planning and Scheduling, Nancy, France, October 26-30, 2020}, pages
  101--109. {AAAI} Press, 2020.
\newblock URL \url{https://aaai.org/ojs/index.php/ICAPS/article/view/6650}.

\bibitem[Fabiano et~al.(2021)Fabiano, Burigana, Dovier, Pontelli, and
  Son]{DBLP:conf/pricai/FabianoBDPS21}
Francesco Fabiano, Alessandro Burigana, Agostino Dovier, Enrico Pontelli, and
  Tran~Cao Son.
\newblock Multi-agent epistemic planning with inconsistent beliefs, trust and
  lies.
\newblock In \emph{{PRICAI} 2021, Hanoi, Vietnam, November 8-12, 2021,
  Proceedings, Part {I}}, volume 13031 of \emph{Lecture Notes in Computer
  Science}, pages 586--597. Springer, 2021.
\newblock \doi{10.1007/978-3-030-89188-6\_44}.

\bibitem[Bolander et~al.(2015)Bolander, Jensen, and
  Schwarzentruber]{bolander2015complexity}
T.~Bolander, M.H. Jensen, and F.~Schwarzentruber.
\newblock Complexity results in epistemic planning.
\newblock In \emph{{IJCAI} International Joint Conference on Artificial
  Intelligence}, volume 2015-January, pages 2791--2797, 2015.

\bibitem[Rinartha et~al.(2018)Rinartha, Suryasa, and Kartika]{8692996}
Komang Rinartha, Wayan Suryasa, and Luh Gede~Surya Kartika.
\newblock Comparative analysis of string similarity on dynamic query
  suggestions.
\newblock In \emph{2018 Electrical Power, Electronics, Communications, Controls
  and Informatics Seminar (EECCIS)}, pages 399--404, 2018.
\newblock \doi{10.1109/EECCIS.2018.8692996}.

\bibitem[Wang et~al.(2021)Wang, Wang, Joty, and Hoi]{wang2021codet5}
Yue Wang, Weishi Wang, Shafiq Joty, and Steven~CH Hoi.
\newblock Codet5: Identifier-aware unified pre-trained encoder-decoder models
  for code understanding and generation.
\newblock \emph{arXiv preprint arXiv:2109.00859}, 2021.

\bibitem[Gigerenzer and Brighton(2009)]{gigerenzer2009}
G.~Gigerenzer and H.~Brighton.
\newblock {{H}omo heuristicus: why biased minds make better inferences}.
\newblock \emph{Top Cogn Sci}, 1\penalty0 (1):\penalty0 107--143, Jan 2009.
\newblock \doi{10.1111/j.1756-8765.2008.01006.x}.

\bibitem[Helmert(2006)]{helmert2006fast}
Malte Helmert.
\newblock The fast downward planning system.
\newblock \emph{Journal of Artificial Intelligence Research}, 26:\penalty0
  191--246, 2006.

\bibitem[Cox(2005)]{Cox2005}
Michael~T. Cox.
\newblock Metacognition in computation: A selected research review.
\newblock \emph{Artificial Intelligence}, 169\penalty0 (2):\penalty0 104--141,
  2005.
\newblock ISSN 0004-3702.
\newblock \doi{https://doi.org/10.1016/j.artint.2005.10.009}.
\newblock URL
  \url{https://www.sciencedirect.com/science/article/pii/S0004370205001530}.
\newblock Special Review Issue.

\bibitem[Kralik et~al.(2018)Kralik, Lee, Rosenbloom, Jackson, Epstein, Romero,
  Sanz, Larue, Schmidtke, Lee, and McGreggor]{Kralik2018}
Jerald~D. Kralik, Jee~Hang Lee, Paul~S. Rosenbloom, Philip~C. Jackson, Susan~L.
  Epstein, Oscar~J. Romero, Ricardo Sanz, Othalia Larue, Hedda~R. Schmidtke,
  Sang~Wan Lee, and Keith McGreggor.
\newblock Metacognition for a common model of cognition.
\newblock \emph{Procedia Computer Science}, 145:\penalty0 730--739, 2018.
\newblock ISSN 1877-0509.
\newblock \doi{https://doi.org/10.1016/j.procs.2018.11.046}.
\newblock URL
  \url{https://www.sciencedirect.com/science/article/pii/S1877050918323329}.
\newblock Postproceedings of the 9th Annual International Conference on
  Biologically Inspired Cognitive Architectures, BICA 2018 (Ninth Annual
  Meeting of the BICA Society), held August 22-24, 2018 in Prague, Czech
  Republic.

\bibitem[Posner(2020)]{Posner2020}
Ingmar Posner.
\newblock Robots thinking fast and slow: On dual process theory and
  metacognition in embodied {AI}, 2020.
\newblock URL \url{https://openreview.net/forum?id=iFQJmvUect9}.

\bibitem[Kerschke et~al.(2019)Kerschke, Hoos, Neumann, and
  Trautmann]{DBLP:journals/ec/KerschkeHNT19}
Pascal Kerschke, Holger~H. Hoos, Frank Neumann, and Heike Trautmann.
\newblock Automated algorithm selection: Survey and perspectives.
\newblock \emph{Evol. Comput.}, 27\penalty0 (1):\penalty0 3--45, 2019.
\newblock \doi{10.1162/evco\_a\_00242}.

\bibitem[Shenhav et~al.(2013)Shenhav, Botvinick, and Cohen]{Shenhav2013}
Amitai Shenhav, Matthew~M. Botvinick, and Jonathan~D. Cohen.
\newblock The expected value of control: An integrative theory of anterior
  cingulate cortex function.
\newblock \emph{Neuron}, 79\penalty0 (2):\penalty0 217--240, July 2013.
\newblock ISSN 0896-6273.
\newblock \doi{10.1016/j.neuron.2013.07.007}.

\bibitem[Haldar and Mukhopadhyay(2011)]{haldar2011levenshtein}
Rishin Haldar and Debajyoti Mukhopadhyay.
\newblock Levenshtein distance technique in dictionary lookup methods: An
  improved approach.
\newblock \emph{arXiv preprint arXiv:1101.1232}, 2011.

\bibitem[Gupta et~al.(1991)Gupta, Nau, et~al.]{gupta1991complexity}
Naresh Gupta, Dana~S Nau, et~al.
\newblock Complexity results for blocks-world planning.
\newblock In \emph{AAAI}, volume~91, pages 629--633. Citeseer, 1991.

\bibitem[Kominis and Geffner(2015)]{kominis2015beliefs}
Filippos Kominis and Hector Geffner.
\newblock Beliefs in multiagent planning: From one agent to many.
\newblock In \emph{Proceedings of the Twenty-Fifth International Conference on
  Automated Planning and Scheduling, {ICAPS} 2015, Jerusalem, Israel, June
  7-11, 2015}, pages 147--155. {AAAI} Press, 2015.
\newblock URL
  \url{http://www.aaai.org/ocs/index.php/ICAPS/ICAPS15/paper/view/10617}.

\bibitem[Huang et~al.(2017)Huang, Fang, Wan, and Liu]{huang2017general}
Xiao Huang, Biqing Fang, Hai Wan, and Yongmei Liu.
\newblock A general multi-agent epistemic planner based on higher-order belief
  change.
\newblock In \emph{Proceedings of the Twenty-Sixth International Joint
  Conference on Artificial Intelligence, {IJCAI} 2017, Melbourne, Australia,
  August 19-25, 2017}, pages 1093--1101. ijcai.org, 2017.
\newblock \doi{10.24963/ijcai.2017/152}.

\bibitem[Muise et~al.(2015)Muise, Belle, Felli, McIlraith, Miller, Pearce, and
  Sonenberg]{muise2015planning}
Christian~J. Muise, Vaishak Belle, Paolo Felli, Sheila~A. McIlraith, Tim
  Miller, Adrian~R. Pearce, and Liz Sonenberg.
\newblock Planning over multi-agent epistemic states: {A} classical planning
  approach.
\newblock In \emph{Proceedings of the Twenty-Ninth {AAAI} Conference on
  Artificial Intelligence, January 25-30, 2015, Austin, Texas, {USA}}, pages
  3327--3334. {AAAI} Press, 2015.
\newblock URL
  \url{http://www.aaai.org/ocs/index.php/AAAI/AAAI15/paper/view/9974}.

\end{thebibliography}

\end{document}